# Multi-target normal behaviour models for wind farm condition monitoring

Angela Meyer*

***Abstract—*** **The trend towards larger wind turbines and remote locations of wind farms fuels the demand for automated condition monitoring strategies that can reduce the operating cost and avoid unplanned downtime. Normal behaviour modelling has been introduced to detect anomalous deviations from normal operation based on the turbine's SCADA data. A growing number of machine learning models of the normal behaviour of turbine subsystems are being developed by wind farm managers to this end. However, these models need to be kept track of, be maintained and require frequent updates. This research explores multi-target models as a new approach to capturing a wind turbine's normal behaviour. We present an overview of multi-target regression methods, motivate their application and benefits in SCADA-based wind turbine condition monitoring, and assess their performance in a wind farm case study. We find that multi-target models are advantageous in comparison to single-target modelling in that they can reduce the cost and effort of practical condition monitoring without compromising on the accuracy. We also outline some areas of future research.**

## 1. Introduction

W IND power forms a major source of renewable energy in our energy systems. The operating cost of wind farms have remained relatively constant over the past decade, making up one quarter of the lifetime cost of onshore farms and one third of the offshore farm cost [1]. As the levelized cost of wind energy keeps falling, the relative importance of its operating cost continues to grow. Automated sensor-based condition monitoring provides powerful capabilities for detecting and diagnosing faults in wind turbines at an early stage of their development. Thereby, it facilitates informed operation and maintenance decision making of wind farm asset managers and cost-efficient preventive measures, in particular the scheduling and performance of condition-based maintenance actions.

The operating state and conditions of modern multi-MW wind turbines are being monitored with various sensor systems [2][3]. Most of the modern wind turbines are equipped with a supervisory control and data acquisition (SCADA) system which logs a multitude of operating and environmental state variables at regular intervals of 1 to 15 minutes, such as the wind speed, the generated active power and the rotor speed. Additionally, the tower, nacelle and drive train components are often equipped with vibration sensors which monitor the radial and axial vibrational motion of housings, bearings, gearbox and generator shafts at kHz sampling rates. Many large turbines comprise additional monitoring systems, for instance oil quality monitoring systems that are integrated with the gear box.

Today's automated wind farm monitoring and fault detection is facing several challenges as the fleet sizes and the abundance of operating data are growing. A major challenge comes from the large variety of fault types that are potentially observable in a wind farm portfolio and the low numbers of actually observed instances for each fault type. This lack of observed faults is likely to result in strong class imbalance and potentially a lack of representativeness if supervised machine-learning based fault detection and diagnosis methods were being pursued. So instead, it is often more feasible to learn a representation of the turbine's normal behaviour and detect relevant deviations from this behaviour. Note that the normal behaviour may change over time due to events like software updates, sensor recalibrations or part replacements, or due to slow processes for instance in relation to normal ageing.

A second challenge is related to the large volume and variety of data collected from hundreds of sensor systems in every turbine. Methods are required that can efficiently analyse the collected data and also deal with different sampling rates ranging from sub-Hertz to kHz rates. In practice, wind farm managers have started to develop a growing number of statistical and machine learning models that describe the normal behaviour of specific turbine subsystems. This may include, for instance, a model of the temperature behaviour of a main bearing component, a second model of the generated active power, and so forth. All of these models need to be kept track of, be maintained and updated when required. The same applies to any related threshold values for detecting deviations from the normal operating behaviour. Therefore, each additional normal behaviour model increases the effort for the asset managers. This study is the first, to the best of our knowledge, to discuss this challenge and to propose and benchmark multi-target normal behaviour models for addressing them for low-frequency SCADA system data.

A growing body of literature studies and compares data-driven methods for monitoring wind turbines and their subsystems based on models of the turbines' normal operating behaviour. These methods rely on modelling the single target variable of interest as a function of input variables like the wind speed. Single-target models constitute the state of the art in the SCADA-based condition monitoring of wind turbines [4], [5]. For instance, single-target models of the turbine's active power generation were proposed and studied in [6]-[14] for monitoring the operating performance. Single-target models of component temperatures have also become common in fault detection applications [15]-[22]. For instance, models of the generator temperature [17] and of the gearbox temperature [16], [21] were presented for the detection of incipient fault processes in these subsystems. A normal behaviour model of the gearbox lubricant pressure was developed in [22]. More recently, comprehensive reviews of SCADA-based normal behaviour modelling of wind turbines have been provided [4], [5].

* Prof. Dr. Angela Meyer leads a research group in applied machine learning and industrial analytics at the Bern University of Applied Sciences, Switzerland (e-mail: angela.meyer@bfh.ch).

The literature has focused on modelling the normal behaviour of single target quantities. However, it is desirable for asset managers to monitor as many critical turbine subsystems as is practically feasible and cost-efficient for them – and to do so for all of the turbine types and configurations in their portfolio. Especially in large portfolios, but even for single wind turbines, the state-of-the-art single-target monitoring approach quickly results in hundreds and thousands of separate single-target models and threshold values, which need to be maintained, updated and kept track of. This considerable practical challenge has neither been discussed nor addressed in the literature so far. The present study aims to close this gap by introducing multi-target machine-learning models of the turbine normal behaviour and comparing their performance to single-target models and across model types based on minute-scale SCADA data.

Multi-target models are established techniques in the fields of applied statistics and machine learning [23]-[25]. Their application in wind turbine (WT) condition monitoring is novel in view of the state of research in this field [4], [5]. Thus, the aim of this work is to study and benchmark multi-target machine learning models as a new approach to capturing a wind turbine's normal behaviour. A low prediction error in times of normal behaviour is a necessary requirement of normal behaviour models. A focus of this study is to evaluate and compare the prediction errors of different model architectures in times of normal operation. We benchmark six multi-target regression models trained with deep neural networks and with classical machine learning algorithms on the multivariate SCADA time series data provided by nine onshore wind turbines situated in continental Europe. The performance of the multi-target models is then compared to that of common single-target turbine models. We investigate if the multi-target models' predictive performance improves by taking also past observations into account in addition to same-time observations. Moreover, we study whether deep neural networks outperform other machine learning models based on the SCADA observations.

This paper is organized as follows. Section 2 describes the state of research in normal behaviour modelling for WT condition monitoring. Section 3 introduces multi-target models and how an example single-target regression algorithm can be generalized for multiple outputs. The data sources, algorithms, model training and testing are outlined in section 4. Section 5 presents the results of the analysis. Conclusions and possible future work are presented in the final section 6.

## 2. Normal behaviour modelling in wind turbine condition monitoring

Normal behaviour models describe the normal fault-free operation behaviour that is expected under the observed operating and environmental conditions. Given these conditions, normal behaviour models describe the state of a particular subsystem, such as the temperature of a drive train component. The residuals of the expected and the actually measured state of the component are computed to identify unusual operating states. Deviations from the normal behaviour including underperformance and faults can then be detected based on the statistical distribution of the residuals. Deviations can have a multitude of causes, including component operation faults, component damage, control faults, incorrect control settings, and sensor malfunction. Normal behaviour models have been built for active power monitoring [6]-[14] and for various subsystems and components of wind turbines. For instance, the operating behaviour of the generator temperature [17], [27], the gearbox temperature [16], [21], the gearbox lubricant pressure [22] and the tower top motion [28] have been modelled for the detection of incipient fault processes in these subsystems. The authors of [11] presented a wind turbine monitoring system with a comprehensive set of 45 single-target models for monitoring various turbine subsystems. Recent reviews of SCADA-based normal behaviour modelling in wind turbine monitoring are given in [4], [5].

## 3. Multi-target regression models

The usage of characteristic operating curves and single-target regression models in condition monitoring has been presented and demonstrated. In operation, this approach involves repeatedly estimating the parameters of multiple single-target models independently for each model, and then running and maintaining all models concurrently. For a given turbine and wind farm, this approach can lead to a large number of distinct and unrelated models. All of these models and any associated threshold values need to be maintained and kept up to date. In contrast, a multi-target model can predict a large number of target variables simultaneously and thereby significantly reduce the model lifecycle effort.

A turbine's set of normal states can be represented as a hyperplane in a high-dimensional state space. It can be parameterized by the environmental variables, in particular the wind speed at rotor height. If a single input variable is used to parameterize it, the hyperplane takes the shape of a line (Fig. 1). When estimating multiple single-target regression models, the hyperplane (black, Fig. 1) is projected into low-dimensional subspaces of the turbine's state space (light gray and dark gray lines, Fig. 1), and a separate regression model is estimated for each of these subspaces. This results in multiple independent models each of which provides only a partial description of the turbine's normal state.

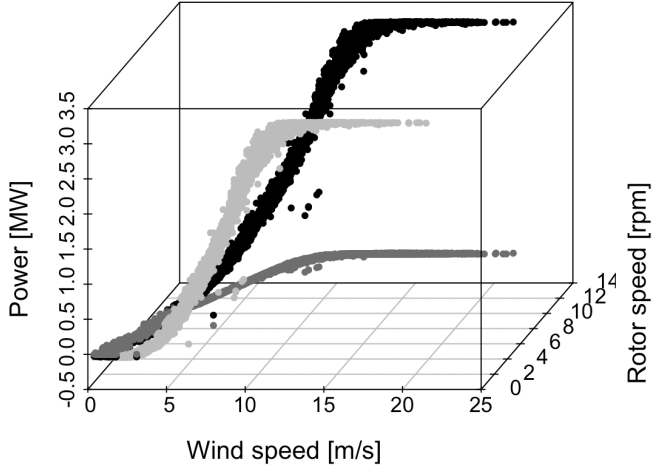

Fig. 1. Turbine state space with projected power and rotor speed curves.

Single-target regression models predict a single continuous real-valued target variable $y$, given at least one input variable $X$. A turbine's active power, for instance, may be estimated as a function of the wind speed or, in a multivariate model, as a function of several input variables such as wind speed, wind direction and air temperature [14].

In contrast, given at least one input variable $X$, a multi-target regression model predicts multiple continuous real-valued target variables $y_1, \ldots y_n$ simultaneously. For instance, a multi-target model may predict both the active power and the rotor speed at once as a function of the wind speed (Fig. 1, black curve). An alternative approach to predicting multiple target variables is to estimate an independent single-target regression model for each target variable. In the previous example, the active power and the rotor speed would each be predicted independently from each other as a function of the wind speed.

| Quantity | Value |
|---|---|
| Rated power | 3.3 MW |
| Hub height | 84 m |
| Rotor diameter | 112 m |
| Cut-in wind speed | 3 m/s |
| Rated wind speed | 13 m/s |
| Cut-out wind speed | 25 m/s |
| Gearbox | 3-stage planetary/helical |

Tab. 1. Technical specifications of the turbine model.

Condition monitoring data from wind farms are subject to phenomena like noise, erroneous sensor signals, numerous correlated input and target variables, and incomplete or missing data. Multi-target regression models may be capable of yielding more accurate predictions in such situations by utilizing the covariance of the target variables according to studies in other fields [30]. Reviews of multi-target learning and specifically multi-target regression algorithms are provided by the authors of [23]-[25]. It has been demonstrated that multi-target models may provide more accurate representations by utilizing the target variables' covariance, and that they may be less prone to overfitting than single-target models [29], [31], [32], [23].

A multi-target regression model is a function approximator that estimates a map $f: \mathbb{R}^k \rightarrow \mathbb{R}^n$ based on a training set $\{(x_{1,i}, \ldots x_{k,i}, y_{1,i}, \ldots y_{n,i}), \ i = 1, \ldots, m\}$ where $m$ is the training set size. The parameters of statistical and machine learning models are estimated by optimizing a loss function. For regression tasks, a common loss function is the sum of squared residuals, $SSR = \sum_{i=1}^{m} \left(y_i^{\text{pred}} - y_i^{\text{obs}}\right)^2$. In a multi-target context, it can be generalized to $SSR = \sum_{i,j} \left(y_{i,j}^{\text{pred}} - y_{i,j}^{\text{obs}}\right)^2$ where $j$ indexes the $n$ target variables $y_1, \ldots y_n$. Typically, the residuals of all $n$ target dimensions are assigned the same weight in the $SSR$, so it can be beneficial to normalize the input variables before training the model.

For illustration, we outline an example of how single-target algorithms have been adapted to multi-target tasks, using the case of decision tree algorithms. Single-target decision trees [33] solve regression tasks by iteratively splitting the training set. Each split is performed so as to maximize the similarity of the data points in the obtained nodes. To this end, the split value in split iteration $l$ is typically estimated by minimizing the weighted sum of squared deviations $J_l^{\text{single}} = \sum_{i,j} \frac{m_j}{m_1+m_2} \left(y_{i,j}^{\text{pred}} - y_j^{\text{mean}}\right)^2$ over the two nodes which result from the split. The nodes are indexed by $j = 1,2$ in this equation, $i$ runs over all $m_j$ elements in node $j$, and $y_j^{\text{mean}}$ denotes the mean value of the regressand variable computed over the elements of node $j$. This single-target algorithm [33] has been generalized to a multi-target one in [34] by summing over all $n$ target dimensions in the loss function, $J_l^{\text{multi}} = \sum_{i,j,p} \frac{m_j}{m_1+m_2} \left(y_{i,j,p}^{\text{pred}} - y_{j,p}^{\text{mean}}\right)^2$ where $p = 1, \ldots, n$.

Generalizations to multi-target predictions have also been accomplished for other regression algorithms including random forests, K nearest neighbours, and artificial neural network algorithms. In the following sections, we are studying and benchmarking multi-target representations for the normal behaviour modelling of wind turbines.

## 4. Model training

### 4.1 Condition monitoring data

One year of SCADA data from nine onshore wind turbines was used to train the multi-target models studied in this work. The results are presented based on one of the turbines' SCADA data and have been confirmed with the SCADA data of eight further turbines. The presented turbine was randomly selected from among the nine available turbines. Its location and the observation time period are not disclosed and the data was anonymized in order to preserve the privacy of the wind turbine owners. All of the nine turbines are a three-bladed variable-speed horizontal-axis model that is gearbox operated and pitch regulated. It is equipped with a SCADA system that provides mean values of operation and environmental state variables at 10-minute intervals. Tab. 1 details its technical specifications.

The 10-minute average wind speed and wind direction measured at nacelle height are utilized as regressors for the model training, validation and testing. The wind speed measured at nacelle height is the most important input variable for predicting the target variables presented in Tab. 2. The wind direction is included to account for any topographic and wake effects upstream of the rotor. The models are trained to predict four target variables whose normal behaviour is affected by the operating conditions: the active power generated by the respective turbine, the rotor speed, the generator speed and the generated current at 10-minute mean values.

| Input variables | Target variables |
|---|---|
| Wind speed [m/s] | Active power [kW] |
| Wind direction [rad] | Rotor speed [rpm] |
| | Generator speed [rpm] |
| | Current [A] |

Tab. 2. Input and target variables.

The wind direction was transformed into a cyclical function by a cosine transformation in order to avoid discontinuities when the wind direction crosses between 360° to 1°. Fig. 2 presents a sample of the input and target data prior to normalization.

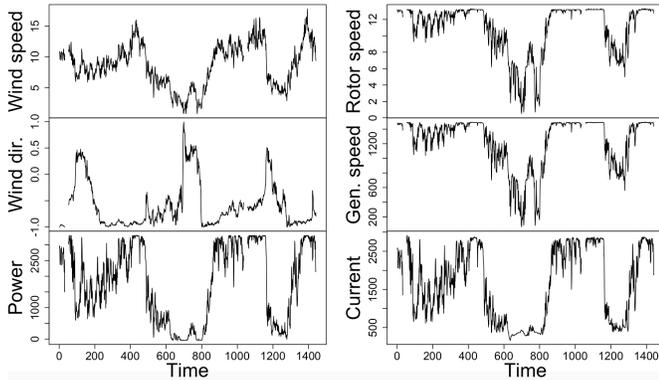

Fig. 2. Input and target variables over a period of ten days of the training set. Wind direction is provided dimensionless after cosine transformation. The units of the remaining variables are provided in Table 2.

### 4.2 Algorithms and model architectures

The multi-target models take the form $y_1, \ldots y_4 \sim x_1 + x_2$ wherein $x_1, x_2$ are the wind speed and direction, and where $y_1, \ldots y_4$ are the active power, rotor and generator rotational velocity and current, respectively. The single-target models have the form $y_i \sim x_1 + x_2, i = 1, \ldots, 4$. The predictive performance of the native multi-target regression models shown in Tab. 3 was studied. The models comprise standard densely connected feed-forward neural networks also called multilayer perceptrons (MLP), convolutional neural networks (CNN) and long short-term memory networks (LSTM). MLPs and CNNs are feed-forward neural networks which feed and transform data in one direction from input to target variables. The CNNs are composed of convolutional feature extraction layers followed by a fully connected multi-target regressor network. Unlike MLPs, CNNs filter and pool the input data to distill features from it before passing on the transformed data to the fully connected layers for training. The filtering is performed with a set of convolutional filters. CNNs do not require feature engineering and can learn time invariant features by iterative convolution and backpropagation.

| Model | ARCHITECTURE |
|---|---|
| Decision tree [33] | Maximum tree depth is 7, minimum number of samples at split is 105, minimum number of samples in leaf is 28. |
| Random forest [35] | Forest of 150 trees, maximum tree depth is 7, minimum number of samples at split is 120, minimum number of samples in leaf is 30. |
| KNN [36] | K=45 nearest neighbours. |
| MLP [37] | Three dense hidden layers with 17, 8, and 17 neurons, a 4-node output layer. |
| CNN [38] | A convolutional layer with 16 filters and kernel length of 5, followed by a max pooling layer with window size 2 and a second convolutional layer with 16 filters, followed by a dense layer of 10 fully connected nodes and a 4-node output layer. |
| LSTM [39] | Three hidden layers with 16, 32, and 16 neurons, respectively. |

Tab. 3. Multi-target model architectures.

CNNs and LSTMs make predictions based on a sequence of past and present observations, whereas the tree-based, KNN and the MLP models predict the target time series values $y_{1,i}, \ldots y_{n,i}$ based on same-time input $x_{1,i}, \ldots x_{k,i}$. In this study, the input time series are fed to the CNN from a trailing window. LSTMs are a recurrent neural network architecture that accepts sequences of input data and passes the data internally from one time step to the next. Random forest is a tree-based ensemble algorithm in which individual trees are combined to achieve a higher predictive performance for the resulting model. The KNN regression predicts the target variables based on the target values of the nearest neighbours inputs.

| Model | ARCHITECTURE |
|---|---|
| Decision tree [33] | Maximum tree depth is 9, minimum number of samples at split is 80, minimum number of samples in leaf is 28. |
| Random forest [35] | Forest of 150 trees, maximum tree depth is 9, minimum number of samples at split is 90, minimum number of samples in leaf is 30. |
| KNN [36] | K=33 nearest neighbours. |
| MLP [37] | Three dense hidden layers with 17, 14, and 17 neurons. |
| CNN [38] | A convolutional layer with 16 filters and kernel length of 5, followed by a max pooling layer with window size 2 and a second convolutional layer with 16 filters, followed by a dense layer of 16 fully connected nodes. |
| LSTM [39] | Three hidden layers with 32 neurons each. |

Tab. 4. Single-target model architectures.

A common modelling framework was developed for this study in order to enable a fair comparison of the algorithms. Six regression model types were trained and tested for single and multiple target variables. The resulting final architectures and hyperparameters are summarized in Tables 3 and 4. The model architectures were built and adapted so as to achieve high training and validation set accuracies. The hyperparameter selection was performed based on a grid search approach for all models except the artificial neural networks (ANNs). For the ANNs, a heuristic hyperparameter selection process was employed by iteratively setting the number of layers and neurons in each layer to obtain a low prediction error on the validation set. The model complexity and thus the number of tunable model parameters was increased only if this provided an increase in prediction accuracy.

For both the multi- and the single-target MLPs, the training process involved the following steps: First, models with a single hidden layer were trained with the number of hidden nodes between 2 and 20. We found that higher accuracies (RMSE) on the training and validation sets could be achieved with two hidden layers rather than one. During the training, the convergence was monitored in terms of the reduction in validation loss. The RMSE was computed for all MLPs with two hidden layers with $s$ nodes in the first layer and $p$ nodes in the second hidden layer for all combinations $(s, p)$ with $s, p = 2, \dots, 20$. The two models with the lowest RMSE were selected from among the resulting models, and a third hidden layer with $q$ nodes was added to each of them. For these three-layer models, the RMSE was computed for all $q = 2, \dots, 20$ nodes. The resulting three-layer model m* with the lowest RMSE was selected. A fourth layer with $t = 2, \dots, 20$ hidden nodes was added but this did not increase the accuracy any further. Thus, the three-layer model m* was chosen as the final model, detailed in Tables 3 and 4. For the multi-target MLPs, the RMSE was computed as the sum of residuals over all four target variables during the training and validation. The training process was stopped early when no increase in accuracy was achieved on the validation set for three training epochs in a row. The training process converged to reach maximum validation set accuracy within 10 to 39 training epochs for the ANNs. To facilitate a robust and fast hyperparameter search, the hidden unit values of the ANNs were normalized by batch normalization [40]. The Adaptive Moment Estimation (Adam) optimizer was applied [41]. The training of the multi- and single-target CNNs started from a first convolutional layer of $k = 8, 16, 32, 64$ filters. The validation set accuracy increased when a second convolutional layer was added. The accuracy of the CNN with two convolutional layers was computed for all combinations of $(k, m)$ with $k, m = 8, 16, 32, 64$ filters in the first and second convolutional layers. Adding a third convolutional layer did not increase the accuracy further. A corresponding strategy was adopted for the LSTM hyperparameter optimization. The optimal number of filters, kernel sizes and input lengths for the CNN and LSTM was obtained by a grid search optimization in which models were estimated in turn with parameter combinations along the grid spanned by the parameters. The parameter combination that resulted in the highest validation-set accuracy was selected for the final model. A grid search optimization was also performed in the case of the random forest algorithms for the number of trees in the forest, the maximum depth of the trees, the minimum number of observations needed to split a node, and the minimum number of observations in each leaf. In the case of the KNN algorithm, KNN models were fit and validated for all odd numbers of nearest neighbours up to 101.

## 5. MODEL ANALYSIS

All input and target variables were normalized before training the models. This facilitated rendering the loss function more symmetric and thereby enable a fast minimization. We normalized the target variables to prevent the predictions from getting dominated by those target variables which take the largest absolute values, as this would be at the expense of the remaining target variables' prediction accuracy. The turbine data was split into training, validation and test sets with a 60% - 20% - 20% split ratio.

For the CNNs we found that 16 filters in the first two convolutional layers resulted in the highest validation set accuracy and outperformed models with even larger numbers of filters. Adding more convolutional layers did not improve the validation set accuracy. The accuracy was found to be weakly sensitive to the convolutional kernel size. A kernel window of size 5 was found optimal. The prediction accuracy also weakly depended on the length of the time series sequence input to the first convolutional layer. An input length of 20 time steps resulted in the highest validation set prediction accuracy. So the CNN's predictions of $y_{1,i}, \dots y_{n,i}$ were most accurate when the CNN was trained on a history of 3 hours of SCADA data $\{(x_{1,i}, \dots x_{k,i}), i = t - 19, \dots, t\}$. The same length of input training data sequences also provided the most accurate predictions in the case of the LSTM.

One might expect that the multi-target models usually have more complex architectures than the single-target models. We found that this is not always the case. For instance, the MLP's second hidden layer comprises 8 neurons in the multi-target MLP and 14 neurons in the single-target case. This might be due to the correlation of the target variables which is leveraged so that fewer parameters suffice to approximate the mapping at highest possible overall accuracy. However, a clear attribution is difficult to provide due to the black box nature of the model.

To test the performance of the normal behaviour models, the mean squared error computed from the predicted and actual target time series was evaluated on the test set. A multi-target model's accuracy of predicting a particular target variable $y_i$ can be expressed by the mean squared error, or likewise its root (RMSE), of the prediction $y_i^{pred}$ with regard to the observed instances $y_i^{obs}$ of the target variable, $MSE(y_i^{pred}, y_i^{obs}) = \frac{1}{s} \sum_{i=1}^{s} (y_i^{pred} - y_i^{obs})^2$, where $s$ denotes the size of the test set. A multi-target model's global root mean squared error is the mean squared error over all instances of all target variables. The six models featured similar levels of prediction accuracy on the test set. As shown in Table 5, the tree-based and neighbour-based models exhibited the highest prediction accuracies from among the six model types. The neural network models produced somewhat larger but acceptable global RMSEs in describing the turbine's normal behaviour. With regard to individual target variables, the rotor speed and the generator speed were associated with the lowest prediction errors by all six models.

| Model | Global RMSE | Power | Rotor speed | Generator speed | Current |
|---|---|---|---|---|---|
| Decision tree | 0.13 | 0.13 | 0.10 | 0.10 | 0.18 |
| Random forest | 0.13 | 0.13 | 0.10 | 0.10 | 0.17 |
| KNN | 0.13 | 0.13 | 0.11 | 0.10 | 0.18 |
| MLP | 0.14 | 0.15 | 0.11 | 0.11 | 0.19 |
| CNN | 0.15 | 0.16 | 0.12 | 0.11 | 0.20 |
| LSTM | 0.14 | 0.13 | 0.11 | 0.11 | 0.18 |

Tab. 5. Accuracies of the multi-target models on the test set.

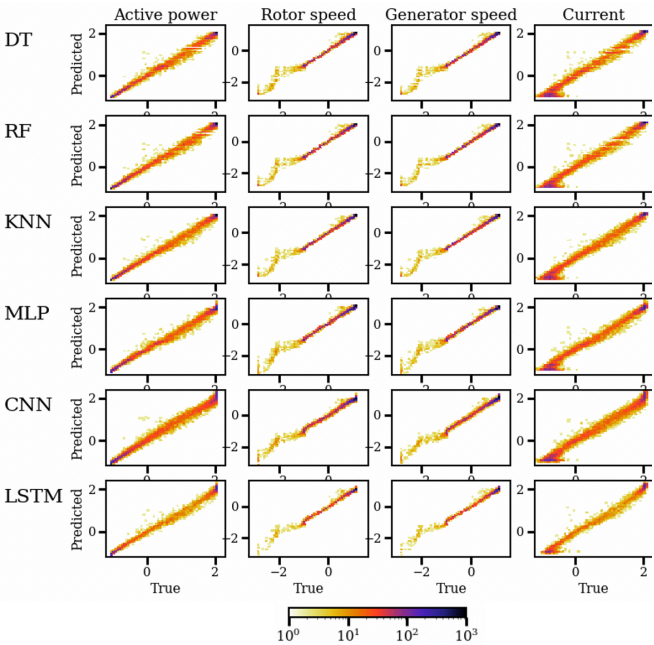

Fig. 3. Distributions of predicted and observed values of the normalized target variables for the test dataset. Each row shows the four target variables (active power, rotor speed, generator speed, current) as observed and as predicted by one of the six multi-target models. Decision tree (DT), random forest (RF), KNN, MLP, CNN and LSTM with model architectures provided in Table 3. The colors indicate the number of observations (dimensionless) in the respective bins of the 2D histograms.

While the ANN models performed somewhat worse than the tree-based and KNN models, the convolutional neural network exhibited the largest prediction errors amongst the models in this case study. Fig. 3 illustrates the predicted and observed values of the normalized target variables. It is confirmed that all six normal behaviour models perform reasonably accurately and that they demonstrate comparable predictive performance. As seen in Fig. 3, most of the predicted and observed values lie on the 45° line.

In Fig. 3, some lower rotor and generator speed values exhibit increased prediction errors. These may possibly be further reduced by considering additional explanatory variables or even by oversampling from the lower rotor and generator speed values as they are underrepresented in the left-skewed distributions of rotor and generator speeds. However, this is not the focus of the present study.

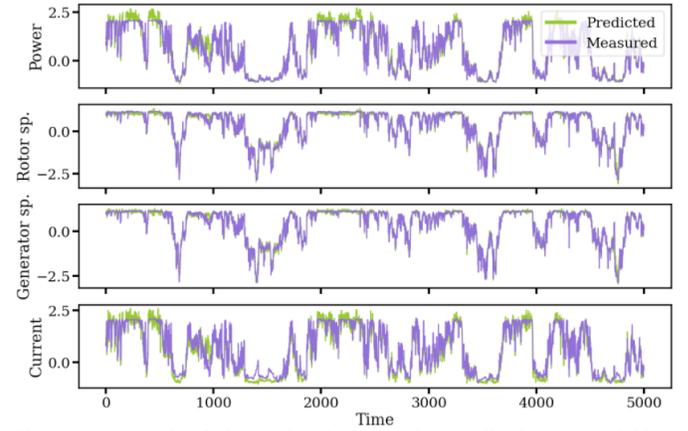

Fig. 4. One month of observed and predicted normalized target variables. Unlike the tree-based models, the CNN overestimates the peak power and current values. The variables are dimensionless due to normalization.

The tree-based and KNN models demonstrated the highest accuracy among all trained normal behaviour multi-target models. In alignment with the low RMSE values, the quantile-quantile comparison of predicted and observed target values indicated a high goodness of fit on the test set for each of the six models. Unlike the other models, the ANNs exhibited some difficulty in accurately predicting tails of the distributions. They tended to somewhat overestimated the upper tails of the power and current distributions, as indicated in Figs. 3 and 4 for the case of the CNN.

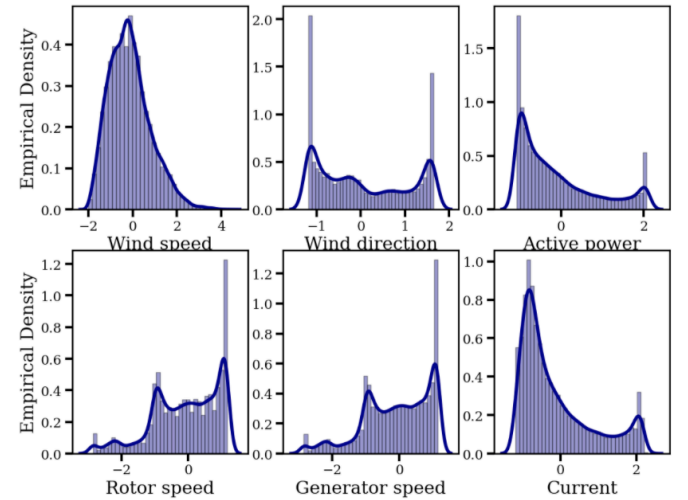

Fig. 5. Density distributions of input and target variables in the training set. The variables are dimensionless due to normalization.

These difficulties may be related to the fact that the training data is not balanced, as illustrated in Fig. 5, which is due to the distribution of wind speeds and wind directions as observed at the turbine site. Regression models that have been trained to minimize the overall error, like the artificial neural network models, can be sensitive to unbalanced data.

Finally, we compared the performance of the trained multi-target models to that of the single-target models, which are more commonly used for modelling the normal behaviour of wind turbines. To this end, the active power $y_1$ was arbitrarily selected from the set of four target variables. The architectures of the trained active power models $y_1 \sim x_1 + x_2$ that provided the highest prediction accuracies are described in Table 4. It was found that the multi-target models provided the same level of prediction accuracy as the single-target models, as illustrated in Tab. 6. This makes multi-target models a convenient alternative for reducing the number of models required to describe a turbine's normal behaviour and for thereby lowering the associated model lifecycle management effort, without compromising on the representation accuracy.

| Model | RMSE |
|---|---|
| Decision tree | 0.13 |
| Random forest | 0.13 |
| KNN | 0.13 |
| MLP | 0.14 |
| CNN | 0.16 |
| LSTM | 0.13 |

Tab. 6. Accuracy of the single-target power model on the test set.

| Model | Multi-target | Single-target |
|---|---|---|
| Decision tree | 0.05 | 0.05 |
| Random forest | 4.0 | 4.1 |
| KNN | 0.02 | 0.02 |
| MLP | 4.9 (23) | 4.5 (16) |
| CNN | 24.9 (34) | 27.5 (39) |
| LSTM | 55.8 (10) | 108.4 (18) |

Tab. 7. Time in seconds required for training a model on an AMD EPYC 7B12 2.25 GHz processor. The number of training epochs is provided in parentheses for the neural network models. It indicates the number of epochs after which the training is stopped as the model performance stopped improving three times in a row on the validation set.

We also compared the computing time required for training the models described in Tables 3 and 4. The training was performed on an AMD EPYC 7B12 2.25 GHz processor. As shown in Table 7, training regression models with multiple target variables did not create a significant increase in model training time in the case of highly covariant target variables while an increase in training time was observed for four of six models in a low-covariance case presented below. In all cases, the neural network models required the longest training times, the LSTM models being the most compute-intensive.

The proposed multi-target approach does not require the target variables to be independent from each other. In this case study, the rotor speed and the generator speed were selected to be among the four target variables. They are closely coupled via the gearbox ratio and not linearly independent. We find that the multi-target models predict both the rotor speed and the generator speed with similar and relatively high accuracy (Fig. 3). In doing so, the models implicitly estimate the gearbox ratio from the SCADA data.

In the case discussed above, the target variables (Table 2) are significantly correlated with the wind speed and with each other. We repeated the analysis with a set of target variables that exhibit significantly lower correlation with the wind speed and amongst each other. A normal behavior model of the hydraulic oil temperature was trained to this end based on 10-minute mean wind speed, wind direction and air temperature from nacelle-mounted meteorological sensors. In the multi-target model, the gear bearing temperature and the nacelle temperature were predicted in addition to the hydraulic oil temperature (Table 8). The target variables were chosen from the available SCADA variables for their comparatively low covariance (Pearson correlation coefficients between 0.5 and 0.6). The model architectures, prediction accuracies and the training times of the single- and multi-target models are shown in Tables 9-13. We found also for this case of relatively weakly correlated target variables that the multi-target models can achieve similar accuracy as the respective single-target models in predicting the target variable of interest, in this case the hydraulic oil temperature. The multi-target ANNs even tended to somewhat outperform their single-target counterparts (Tables 11 and 12). This suggests that, starting from a single-target normal operation model, adding additional target variables to the model can even increase the accuracy in the target variable of interest because the multi-target model can exploit correlations among the target variables [46].

| Input variables | Target variables |
|---|---|
| Wind speed [m/s] | Hydraulic oil temperature [K] |
| Wind direction [rad] | Nacelle temperature [K] |
| Air temperature [K] | Gear bearing temperature [K] |

Tab. 8. Input and target variables of the multi-target model in the low-correlation case.

| Model | ARCHITECTURE |
|---|---|
| Decision tree [33] | Maximum tree depth is 12, minimum number of samples at split is 20, minimum number of samples in leaf is 24. |
| Random forest [35] | Forest of 200 trees, maximum tree depth is 12, minimum number of samples at split is 80, minimum number of samples in leaf is 40. |
| KNN [36] | K=11 nearest neighbours. |
| MLP [37] | Two dense hidden layers with 3 and 9 neurons, and a 3-node output layer. |
| CNN [38] | A convolutional layer with 16 filters and kernel length of 5, followed by a max pooling layer with window size 2 and a 4-node output layer. |
| LSTM [39] | Two hidden layers with 16 and 8 neurons, respectively. |

Tab. 9. Multi-target model architectures in the low-correlation case.

A major advantage of the presented data-driven approach over a formula-based monitoring is that the data-driven multi-target approach does not required any knowledge or upfront estimation of the formula parameters (such as the gearbox ratio in the presented case study) and thereby is more easily transferable among different turbine types and configurations. Moreover, unlike formulas, the presented data-driven approach intrinsically accounts for turbine- and site-specific effects such as topographic or wake effects, and can even deal with input from miscalibrated sensors as long as the miscalibration is stable in time.

| Model | Architecture |
|-------|-------------|
| Decision tree [33] | Maximum tree depth is 8, minimum number of samples at split is 20, minimum number of samples in leaf is 24. |
| Random forest [35] | Forest of 100 trees, maximum tree depth is 12, minimum number of samples at split is 8, minimum number of samples in leaf is 40. |
| KNN [36] | K=11 nearest neighbours. |
| MLP [37] | Two dense hidden layers with 5 and 8 neurons. |
| CNN [38] | A convolutional layer with 16 filters and kernel length of 10, followed by a max pooling layer with window size 2, at step size 20. |
| LSTM [39] | One hidden layer with 16 neurons. |

Tab. 10. Single-target model architectures for the hydraulic oil temperature model.

| Model | Global RMSE | Hydraulic oil temp. | Spinner temp. | Gear bearing temperature |
|-------|-------------|--------------------|--------------|-------------------------|
| Decision tree | 0.71 | 0.88 | 0.65 | 0.57 |
| Random forest | 0.67 | 0.82 | 0.63 | 0.52 |
| KNN | 0.66 | 0.8 | 0.6 | 0.58 |
| MLP | 0.60 | 0.71 | 0.53 | 0.54 |
| CNN | 0.53 | 0.7 | 0.48 | 0.34 |
| LSTM | 0.51 | 0.67 | 0.47 | 0.33 |

Tab. 11. Accuracies of the multi-target models on the test set.

| Model | RMSE |
|-------|------|
| Decision tree | 0.83 |
| Random forest | 0.81 |
| KNN | 0.8 |
| MLP | 0.71 |
| CNN | 0.72 |
| LSTM | 0.74 |

Tab. 12. Accuracy of the single-target models of the hydraulic oil temperature on the test set.

| Model | Multi-target | Single-target |
|-------|-------------|--------------|
| Decision tree | 0.049 | 0.041 |
| Random forest | 6.4 | 3.1 |
| KNN | 0.02 | 0.02 |
| MLP | 4.45 | 0.61 |
| CNN | 6.01 | 4.3 |
| LSTM | 57.14 | 19.96 |

Tab. 13. Time in seconds required for training a model on an AMD EPYC 7B12 2.25 GHz processor.

## 6. Conclusions and future work

This study explored the potential of multi-target machine learning models for describing and monitoring the normal operation behaviour of wind turbines based on SCADA data. The proposed approach supports a simultaneous monitoring of multiple state variables at once with a single (multi-target) model. One year of SCADA data from nine onshore turbines located in Europe was used to this end. Neural networks, decision trees and neighbour-based models were analysed. It was found that all investigated model types demonstrated reasonably low prediction errors in times of normal operation, as required for use in performance monitoring or fault detection tasks. The study showed that the models accomplished a similar level of prediction accuracy. The neural network models showed a tendency to overestimate the upper tails of the target distributions for some of the target variables.

The neural network models provided similar predictive performance as the other explored machine learning models. Moreover, the model training and optimization process demonstrated that deep neural networks with a larger number of hidden layers did not outperform the relatively shallow network architectures or the other models of this study. The multi-target models achieved at least the same predictive performance as the more commonly used single-target models. Interestingly, we also found that in the case of strongly correlated target variables the models which take a history of past observations as input – the LSTM and CNN models – did not outperform the models that accept only same-time observations input. Also, their prediction accuracy did not improve by taking a long history of past observations into account. This may be due to the fact that the input and target variables chosen for this study are strongly correlated, so same-time values of 10-minute mean wind speed and wind direction can be sufficient predictors.

The present study indicates that multi-target normal behaviour models are advantageous in comparison to state-of-the-art single-target models. Multi-target methods reduce the number of independent models required to monitor a turbine. Thereby, they facilitate a lower model training and handling effort. This allows to free up time in the wind farm asset management, without compromising on the model accuracy.

Normal behaviour modelling (NBM) of WTs has been studied for more than a decade following its introduction by [42] in 2009. Since then, dozens, and perhaps hundreds, of studies have been published on SCADA-based NBM and its application in SCADA-based fault detection [4], [5]. All of this work has focused on the modelling and monitoring of single state variables, such as the active power generation or the generator temperature.

However, this single-target approach scales linearly with the number of monitored SCADA channels in a real operational setting: If an operator wants to monitor $n$ state variables in one of her WTs, she will need to make use of an ensemble of $n$ single-target models according to the state of the art. Moreover, each single-target model is typically estimated to describe the normal behaviour of one particular state variable for one particular WT. This means that the operator will need to develop $n * m$ separate single-target models in order to monitor her wind farm of $m$ turbines. In contrast, only m models would be required for the same task based on the multi-target approach presented in this study. When used in an operational context, the single-target approach quickly leads to tens of thousands of single-target regression models which all need to be trained, validated, tested, run, updated and looked after in the IT infrastructure of the operator's remote monitoring center. The single-target approach requires more complex automation software to run in her monitoring center. The increased software complexity makes it more error prone and requires higher software testing effort whenever the machine learning automation software is updated. In addition, more person-hours are needed at the center and more storage space is required to build and maintain all of these single-target models.

Two potential advantages of multi-target NBM (that have thus far not been confirmed yet in WT condition monitoring) are an expected higher accuracy of multi-target models and an improved interpretability that have been reported in other fields [30], [43]-[46]. In the second presented case of the hydraulic oil temperature, we indeed found that some of the multi-target models could achieve a somewhat higher prediction accuracy than their single-target counterparts. These advantages were attributed to the multi-target models' capability of capturing dependencies between the target variables. The present study found that multi- and single-target models have a similar accuracy, and did not investigate questions related to model interpretability. The accuracy of NBM is particularly important because it determines the delay with which underperformance and incipient faults can be detected. Therefore, more studies are needed to investigate, in particular, if and under which circumstances multi-target models may provide a higher accuracy in the context of NBM for performance monitoring and for fault detection tasks in WT compared to ensembles of single-target models.

Our results are promising and may serve as a basis for future studies to further elucidate the potential of multi-target models in the condition monitoring of wind turbines. Future work will need to be performed to explore additional research questions in need of further investigation, such as: How can optimal combinations of input and target variables be characterized and identified? What is the potential of multi-target models for fault detection and diagnosis tasks? Under which circumstances can these models achieve prediction accuracies superior to single-target models in wind turbine condition monitoring? How do multi-target models perform for large numbers of target variables, such as hundreds or even thousands of channels retrieved from a SCADA system? How can one characterize the potential of multi-target models in the analysis and modelling of high frequency data, in particular vibration measurements of the turbine drive train and tower top acceleration?

Moreover, more research is needed to identify potential sets of SCADA variables that will benefit from more accurate prediction based on a multi-target prediction as compared to single-target models. A target variable $y$ benefits if its joint prediction with one or multiple other target variables $y'$ results in a higher predictive accuracy of $y$ than a single-target prediction of $y$. It will also be beneficial to investigate which sets of complementary target variables $y'$ will increase the prediction accuracy of $y$ and which sets have no accuracy-increasing effect.

ACKNOWLEDGEMENTS

We thank four anonymous reviewers whose comments helped improve and clarify this manuscript. We also thank Bernhard Brodbeck, Janine Maron, Dimitrios Anagnostos of WinJi AG, Switzerland, and Kaan Duran of Energie Baden-Wuerttemberg EnBW, Germany, for their appreciated collaboration and valuable discussions. This research did not receive any specific grant from funding agencies in the public, commercial, or not-for-profit sectors.